\documentstyle[a4]{article}

\twocolumn
\if@twoside
\else
\fi

\columnwidth \textwidth
\skip\footins 10pt plus 2pt minus 4pt
\newcommand{\ssp}{\setlength{\baselineskip}{9pt}}
\newcommand{\prevex}[1]{(\arabic{examplectr}{#1})}
\newcounter{examplectr}
\newcounter{subexamplectr}
\newenvironment{subex}%
   { \addtocounter{subexamplectr}{1}
     \begin{list}
       {\alph{subexamplectr}}%
       {\setlength{\topsep}{-\parskip}
        \setlength{\leftmargin}{0.15in}
        \setlength{\labelsep}{0.1in}}
       \item \ssp
   }%
   {\end{list}}
\newcommand{\myref}[2]{
     (\arabic{#1}{#2})}
\newcommand{\mylabel}[1]{
   \newcounter{#1}
   \setcounter{#1}{\value{examplectr}}}
\newenvironment{ex}%
   { \addtocounter{examplectr}{1}
     \setcounter{subexamplectr}{0}
     \begin{list}
       {\arabic{examplectr})}%
       {\setlength{\topsep}{0.1in}
        \setlength{\leftmargin}{0.25in}
        \setlength{\labelsep}{0.1in}}
       \item
    }%
    {\end{list}}
\title{\vspace*{-17mm}
{\bf {\footnotesize In Proceedings of the 15th International
Conference on Computational Linguistics, Coling 94, Kyoto, Japan,
pages 935-941.}} \\[3mm]
{\large NON-CONSTITUENT COORDINATION: THEORY AND PRACTICE\thanks{This
research was supported by the U.K. Science and Engineering Research Council
(Research Fellowship B/90/ITF/288).} \\}}
\date{}
\author{{\large David Milward}\\[4mm]
{\normalsize Centre for Cognitive Science, University of Edinburgh}\\
{\small 2, Buccleuch Place, Edinburgh, EH8 9LW, Scotland,
davidm@cogsci.ed.ac.uk}}
\begin{document}
\maketitle
\subsection*{\small {\bf ABSTRACT}}

Despite the large amount of theoretical work done on non-constituent
coordination during the last two decades, many computational systems
still treat coordination using adapted parsing strategies, in a
similar fashion to the SYSCONJ system developed for ATNs. This paper
reviews the theoretical literature, and shows why many of the
theoretical accounts actually have worse coverage than accounts based
on processing.  Finally, it shows how processing accounts can be
described formally and declaratively in terms of Dynamic Grammars.

\subsection*{\small {\bf INTRODUCTION}}

This paper is concerned with {\it symmetrical} coordination, where the
order of the conjuncts (the items being coordinated by a conjunction
such as {\it and} or {\it or}) can be altered without affecting
acceptability.  Coordination of this kind is traditionally split into
{\it constituent coordination}, where each conjunct forms a
constituent according to `standard' phrase structure grammars, and
{\it non-constituent coordination}.  Constituent and non-constituent
coordination have been treated as entirely separate phenomena (see van
Oirsouw, 1987 for discussion), and different mechanisms have been
proposed for each. However, by considering grammaticality judgements
alone, there seems little justification for such a division.  To
illustrate this, consider the sentence:
\begin{ex}
 John gave Mary some books
\end{ex}
Each of the final proper substrings of the
sentence
(i.e.\ {\it some books}, {\it Mary some books} etc.) can be used as
a conjunct e.g.
\begin{ex}
\begin{subex}
John gave Mary [some books] and [some papers]
\end{subex}
\begin{subex}
John gave [Mary some books] and [Peter some papers]
\end{subex}
\begin{subex}
John [gave Mary some books] and [lent Peter some papers]
\end{subex}
\end{ex}
\mylabel{books}
Similarly, each of the initial substrings can be used as a conjunct e.g.
\begin{ex}
\begin{subex}
[John gave] and [Peter lent] Mary some books
\end{subex}
\begin{subex}
[John gave Mary] and [Peter lent George] some books
\end{subex}
\begin{subex}
[John gave Mary some] and [Peter lent George many] books
\end{subex}
\end{ex}
\mylabel{booksi}
and so can each of the middle substrings e.g.
\begin{ex}
\begin{subex}
John [gave Mary some] and [lent Peter many] books
\end{subex}
\begin{subex}
John [gave Mary] and [lent Peter] many books
\end{subex}
\begin{subex}
John gave [Mary some] and [Peter many] books
\end{subex}
\end{ex}
Only examples \myref{books}{a} and \myref{books}{c} are constituent
coordinations.  Example \myref{booksi}{c} seems slightly unnatural,
but it is much improved if we replace {\it books} by a heavier string
such as {\it books about gardening}.  Thus, for this example, any
substring of the sentence\footnote{The examples above only consider
substrings containing more than one word. Coordination of the
individual words (which is necessarily constituent coordination) is
also possible.  Natural examples involving the determiner, {\it some},
are difficult to achieve, however determiner coordination is possible
(consider: {\it I didn't know whether to expect few or many people to
come}).} can form a viable conjunct.

\subsection*{\small {\bf DELETION ACCOUNTS}}

In the last twenty to thirty years there have been a series of
accounts of coordination involving various deletion mechanisms (from
e.g.\ Gleitman, 1965 to van Oirsouw, 1987). For example, from the
following `antecedent' sentence,
\begin{ex}
Sue gave Fred a book by Chomsky and
Sue gave Peter a paper by Chomsky
\end{ex}
van Oir{souw allows deletion of words to the left and to the right
of the conjunction,
{\begin{tabbing}
xxx \= Sue gave \=  Peter a paper \= by Chomsky \kill
\>Sue gave \> Fred a book \> by Chomsky and \\[-5mm]
\> \> \> xx xxxxxxxx \\
\> Sue gave \> Peter a paper \> by Chomsky \\[-5mm]
\> xxx xxxx
\end{tabbing}}
\noindent
resulting in the sentence:
\begin{ex}
Sue gave Fred a book and Peter a paper by Chomsky
\end{ex}
\mylabel{bychomsky} Most deletion accounts assume that deletion is
performed under identity of words, but don't analyse what it means for
two words to be identical (an exception is van Oirsouw who discusses
phonological, morphological and referential identity).  Consider the
following example of deletion.
\begin{ex}
John will drive and Mary built the drive \\
{*} [John will] and [Mary built the] drive
\end{ex}
Here the two cases of {\it drive} are phonologically identical, but
have different syntactic categories.  Now consider:
\begin{ex}
\begin{subex}
{*} John bored [the new hole] and [his fellow workers]
\end{subex}
\begin{subex}
{*} Mary came in [a hurry] and [a taxi]
\end{subex}
\end{ex}
These are cases of `zeugma' and are unacceptable except as jokes.  It
therefore seems that the deleted words must have the same major
syntactic category, and the same lexical meaning.

However, even if we fix both syntactic category and lexical meaning,
we still get some weird coordinations. For example, consider:
\begin{ex}
\begin{subex}
{*} Sue saw$_{i}$ the man$_{j}$ [through the telescope]$_{i}$ and [with the
troublesome kid]$_{j}$
\end{subex}
\begin{subex}
{*}I saw [a friend of] and [the manufacturer of] Mary's handbag
\end{subex}
\end{ex}
\mylabel{handbag}
In example (a) the two prepositions are attached differently, one to
the verb {\it saw}, the other to the noun, {\it man}. In example (b),
attributed to Paul Dekker, the two conjuncts require {\it Mary's
handbag} to have a different syntactic structure: the bracketing
appropriate for the first conjunct is {\it [[a friend of Mary]'s
handbag]}.  The unacceptability of these examples suggests that word by
word identity is insufficient, and that deleted material must have
identical syntactic structure, as well as identical lexical meanings.

Some of the most compelling arguments against deletion have been
semantic. For example, Lakoff and Peters
(1969) argued that deletion accounts are inappropriate for certain
constituent coordinations such as:
\begin{ex}
John and Mary are alike
\end{ex}
since the `antecedent' sentence {\it John are alike and Mary are alike}
is nonsensical (it is also ungrammatical if we consider number agreement).

However, semantically inappropriate or nonsensical `antecedents' are
also possible when we consider non-constituent coordination.  For
example, consider `antecedents' for the following:
\begin{ex}
\begin{subex}
{[}The man who buys] and [the woman who sells] rattlesnakes met outside
\end{subex}
\begin{subex}
Many former [soldiers living in England] and [resistance members
living in France] have similar memories
\end{subex}
\begin{subex}
John sold different dealers [a vase using his intensive sales technique]
and [a bookcase using his market-stall technique]
\end{subex}
\end{ex}
\mylabel{soldiers}
\prevex{b} is non-constituent coordination under the
primary reading where the scope of {\it former} does not contain {\it
living in England} i.e. where the semantic bracketing is:
\begin{ex}
{[}[former soldiers] living in England]
\end{ex}
Examples (a) and (b) could be expanded out at the NP level, but not at
the S level.  However (c) cannot be expanded out at any constituent
level, whilst retaining an appropriate semantics. For example,
expansion at the VP level gives:
\begin{ex}
John sold different dealers a vase using his intensive sales technique
and different dealers a bookcase using his market-stall technique
\end{ex}
Thus, although Lakoff and Peters' arguments count against standard
deletion analyses, they do not count as general arguments against a unified
treatment of constituent and non-constituent coordination.

\subsection*{\small {\bf SHARED STRUCTURE}}

Consider the sentence:
\begin{ex}
John gave Mary a book and Peter a paper by Chomsky
\end{ex}
Instead of thinking of {\it John gave} and {\it by Chomsky}
as deleted, we can also think of them
as shared by the two conjuncts. This structure can be represented
as follows:
{\small
\begin{tabbing}
John gave xx \=  Peter a paper xx \=  \kill
\> Mary a book \\
John gave \> \hspace*{3mm} {\it and} \> by Chomsky \\
\> Peter a paper
\end{tabbing}
}
\noindent
{}From the result of the previous section, each conjunct
must share not just the phonological material, but also
the syntactic structure and the lexical meanings.

There are three main methods by which this sharing of structure can be
achieved: phrasal coordination, 3-D coordination, and processing
strategies.

\subsection*{\small {\bf PHRASAL COORDINATION}}

At first sight, analysing non-constituent coordination using phrasal
(i.e.\ constituent) coordination seems nonsensical. This is not the
case. Coordinations are classified as non-constituent coordination if
the conjuncts fail to be constituents in a `standard' phrase structure
grammar.  However, they may well be constituents in other grammars.
For example, it has been argued that the weaker notion of constituency
provided by Categorial Grammars is exactly what is required to allow
all conjuncts to be treated as constituents (Steedman 1985).

Phrasal coordination is exemplified by the schema \footnote{There have
been various arguments (stemming from Ross 1967) for the adoption of
a variant of this schema, in which the coordinating conjunctions is
associated solely with the last conjunct.  The schema is revised as
follows:
\begin{quote}
X \hspace{2mm} $\rightarrow$ \hspace{2mm} X \hspace{2mm}
X[Conj] \\[2mm]
X[Conj] \hspace{1mm} $\rightarrow$ Conj X
\end{quote}
}:
\begin{quote}
X \hspace{2mm} $\rightarrow$ \hspace{2mm} X \hspace{1mm} Conj \hspace{1mm}
X
\end{quote}
The shared material is necessarily treated identically for each
conjunct since there is only a single copy: the conjunction is
embedded in a single syntax tree.

The phrasal coordination schema requires each conjunct to be given a
single type, and for the conjuncts and the conjunction as a whole to
be of the same type. Problems with the latter requirement were pointed
out by Sag et al.\ (1985), who gave the following counterexamples:
\begin{ex}
\begin{subex}
We walked slowly and with great care
\end{subex}
\begin{subex}
Pat is a Republican and proud of it
\end{subex}
\begin{subex}
I am hoping to get an invitation and optimistic about my chances
\end{subex}
\end{ex}
\mylabel{sag} Sag et al. deal with these examples by treating
categories as feature bundles, and allowing coordination in cases
where there are features in common.  For example, the two conjuncts in
\myref{sag}{a} share the feature {\small {\it $+$MANNER}}\footnote{Sag
  et al. also suggest an alternative treatment using an apparently
  otherwise unmotivated grammar rule AdvP $\rightarrow$ PP.}. As it
stands, the account does not deal with examples such as the following,
\begin{ex}
TNT deliver efficiently and on Sundays
\end{ex}
Here the adverbial phrase would presumably be {\small {\it
$+$MANNER}}, and the prepositional phrase, {\small {\it $+$TEMP}}.
Further examples which are problematic for Sag et al.\ are given by
Jorgensen and Abeill\'{e}, (1992).

An alternative, suggested by Morrill (1990) and similar to Jorgensen and
Abeill\'{e} (1992), is to use the following coordination schema:
\begin{quote}
X$\vee$Y \hspace{2mm} $\rightarrow$ \hspace{2mm}
X \hspace{1mm} Conj \hspace{1mm} Y
\end{quote}
This does not impose any condition that the two categories {\bf X} and
{\bf Y} share anything in common. However, the new category {\bf
  X$\vee$Y} is used to ensure that both categories are appropriate in
the context. For example, \myref{sag}{b} is acceptable since the
coordination type is {\bf NP$\vee$AP}, and {\it is} subcategorises for
both NPs and APs.

A rather more difficult problem is that of providing types for all possible
conjuncts. Consider the following:
\begin{ex}
\begin{subex}
Sue gave Fred a book and Peter a paper
\end{subex}
\begin{subex}
Mary admires and Sue thinks she likes Peter
\end{subex}
\end{ex}
\mylabel{gave}
(a) is a conjunction of two pairs of noun phrases.
(b) is a case of `unbounded Right-Node Raising' where the noun phrase
{\it Peter} is embedded at different depths in the two conjuncts.

There have been two main approaches to dealing with examples such as
(a) using phrasal coordination. The first is to introduce an explicit
product operator (e.g.\ Wood 1988), allowing types of the form {\bf
NP$*$NP}.  The second is to use a calculus in which types can undergo
`type-raising' (e.g.\ Dowty 1988), or can be formed by abstraction (as
in the Lambek Calculus, Lambek 1958). The effect is to treat {\it Fred
a book} as a verb phrase missing its verb.

The advantage of adopting a general abstraction mechanism, as in the
Lambek Calculus, is that this also provides a treatment of examples
such as (b).  Unfortunately, the ability to perform abstraction of
categories with functional types (which is required for (a)) also
allows shared material to get different syntactic analyses, resulting
in acceptance of all the sentences predicted by deletion accounts
where identity of lexical categories and lexical semantics is
respected, but not identity of syntactic structure.  Reconsider:
\begin{ex}
{*}I saw [a friend of] and [the manufacturer of] Mary's handbag
\end{ex}
We can obtain identical syntactic types for {\it a friend of} and
{\it the manufacturer of} by subtracting the lexical types of
{\it I}, {\it saw}, {\it Mary}, {\it 's}, and {\it handbag} from the
sentence type S\footnote{The type given to
both conjuncts, using
reasonably standard type assignments and `Lambek' notation, would be:
(((NP$\backslash$(((NP$\backslash$S)/NP)$\backslash$
S))/NP)/(NP$\backslash$(NP/N)))/N}.
Since the types are identical, coordination can then take place. Thus
the ability to `subtract' one type from another allows the Lambek Calculus
to replicate a deletion account, and it thereby suffers from the same
problems.

There have been some proposals to restrict the Lambek Calculus
in order to prevent such overgeneration. Barry and Pickering (1993) propose
a calculus in which \myref{gave}{a} is dealt with using a product operation,
and abstraction is limited to categories which do not act as a function
in the derivation. This account makes reasonably
good empirical predictions, though it does fail for the following examples:
\begin{ex}
\begin{subex}
You can call me directly or after 3pm
through my secretary
\end{subex}
\begin{subex}
Sue put a lamp on the table, and on the ledge a large antique punchbowl
\end{subex}
\end{ex}
\mylabel{secret}
In (a), each conjunct contains different numbers of modifiers of
different types (an adverbial phrase with two prepositional phrases).
In (b) the subcategorisation order is swapped in the two conjuncts.

Successful treatment of non-constituent coordination using phrasal
coordination seems to require elaborate encoding in the conjunct type
of a simple generalisation: conjuncts can coordinate provided they are
acceptable within the same syntactic context.
The 3-D approaches and processing strategies use syntactic context more
directly, and it is to these methods which we now turn.

\subsection*{\small {\bf 3-D Coordination}}

Let us briefly reconsider our explanation of deletion. Example
\myref{bychomsky}{} was explained by saying that the two strings {\it
by Chomsky} and {\it Sue gave} are deleted under some notion of
identity. However, we could equally well have described this as a
process whereby the first instance of {\it by Chomsky} is merged with
the second (under some notion of identity), and the second instance of
{\it Sue gave} is merged with the first.

Merging word strings instead of deleting them does not help with the
problems of deletion accounts which we outlined earlier. In
particular, it does not help to exclude examples \myref{handbag}{a}
and \myref{handbag}{b} which suggest shared material must have
identical syntactic structure. However, once we have started to think
in terms of merging, there is an obvious next step, which is to move
from merging of word strings to merging of syntax trees.  This is the
move made by Goodall (1987), who advocates treating coordination as a
union of phrase markers: ``a `pasting together' one on top of the
other of two trees, with any identical nodes merging together''
(Goodall, 1987, p.20). We can visualise the result in terms of a
three-dimensional tree structure, where the merged material is on one
plane, and the syntax trees for each conjunct are on two other planes.
For example, consider the 3-D tree for example \myref{gave}{a} given
in Fig.~1.
\\
\begin{verbatim}
        s
       / \
      /   \
    np     vp
   Sue     /+ *
          /  +   *
         /   np      np
        /     ++      * *
       /     +  +      *   *
      v'     det  n    det    n
     /+*     a  book    a    paper
    /  +  *
   /    +    *
  v      np     np
gave   Fred     Peter
\end{verbatim}
\hspace*{15mm} {\small {Fig.~1}}
\\[3mm]
The merged part of the tree includes all the nodes which dominate the shared
material {\it Sue gave}. The conjuncts retain separate planes
(denoted here by using stars or crosses for branches).

Goodall's account does not deal with examples such as \myref{gave}{b}, which
he argues to be examples of a different phenomenon. However these can be
incorporated into a 3-D account (e.g.\ Moltmann, 1992).

There are various technical difficulties with Goodall's account (see
e.g.\ van Oirsouw, 1987, and Moltmann, 1992).  There is also a
fundamental problem concerning semantic interpretation of coordinated
structures (see Moltmann, 1992 which provides a revised and more
complex 3-D account based on Muadz, 1991).

For coordination of unlike categories, as in the examples in
\myref{sag}{}, Goodall proposes a treatment somewhat similar to Sag et
al.\ (1985).  However there is still a problem in dealing with examples
where there are different numbers of modifiers, such as \myref{secret}{a}
or the following:
\begin{ex}
\begin{subex}
We can meet at the office or in London outside the theatre
\end{subex}
\begin{subex}
TNT deliver efficiently and after 5pm in Edinburgh
\end{subex}
\end{ex}
Consider example (b). The syntactic structure appropriate for {\it
TNT deliver efficiently} has one S node and two VP nodes.  However,
the structure for {\it TNT deliver after 5pm in Edinburgh} requires
one S node and three VP nodes (or three S nodes and one VP node). The
two structures therefore fail to merge since the structure dominating
the shared material {\it TNT deliver} must be identical.  The use of
ordered phrase structure trees also excludes examples such as
\myref{secret}{b}.

In summary, the 3-D approaches correctly enforce identity of syntactic
structure for shared material. However, the way of characterising
syntactic structure using (parts of) standard phrase structure trees
results in an overly strict requirement of parallelism between the
conjuncts.  We will now consider processing strategies, where syntactic
structure of shared material is characterised more indirectly by the
state of the parser.

\subsection*{\small {\bf PROCESSING STRATEGIES}}

There have been several attempts to treat coordination by adapting
pre-existing parsing strategies. For example, ATNs were adapted by
Woods (1973), DCGs by Dahl and McCord (1983), and chart parsers by
Haugeneder (1992). Woods and Dahl \& McCord's system are similar.
Haugeneder's system has very limited coverage.

In Wood's SYSCONJ system, the parser can back up to various points in
the history of the parse, and parse the second conjunct according to
the configuration found. For example, in parsing,
\begin{ex}
John gave some books to Peter and some papers to George
\end{ex}
at the point after encountering {\it and}, the parser can reaccess the
configuration after parsing {\it John gave} i.e. a stack consisting of
a sentence and a verb-phrase, and an arc traversal by the verb.
The second conjunct is then parsed according to this configuration.

SYSCONJ does not immediately merge the two stack configurations after
completing the second conjunct, but, instead, separately parses both
conjuncts in parallel until a constituent is completed. For example,
on parsing the sentence,
\begin{ex}
John gave Mary a book and Peter a paper about subjacency
\end{ex}
the SYSCONJ system separately parses
{\it Peter a paper about subjacency} and {\it Mary a book about subjacency}
before conjoining at the level of some enclosing constituent (for example
the verb phase). The result is therefore similar to starting with the
sentence:
\begin{ex}
John gave Mary a book about subjacency and gave Peter a paper
about subjacency
\end{ex}
As noted by Dahl and McCord, this mechanism means that SYSCONJ
inherits the problems of nonsensical semantics which plague the deletion
accounts, since {\it John and Mary are alike} is treated the same as
{\it John are alike and Mary are alike}. The mechanism also causes
problems for dealing with nested coordination.  Consider the sentence:
\begin{ex}
John wanted to study medicine when he was eleven, law when he was
twelve, and to study nothing at all when he was eighteen
\end{ex}
The smallest constituent containing {\it to study medicine when he was
eleven} is the verb phase {\it wanted to study medicine when he was
eleven}.  However, if coordination of the first two conjuncts occurs
at this level, it is difficult to see how to deal with the final
conjunct.

Both Woods and Dahl \&  McCord use stack based configurations
rather than a full parsing history. Thus once something is popped off the
stack its internal structure cannot be accessed by the coordination routine.
This rules out examples such as the following,
\begin{ex}
John gave some books to Mary and papers to George
\end{ex}
where the NP, {\it some books} is completed prior to the conjunction being
reached.

Although processing accounts can provide reasonable coverage of the
coordination data, the exact predictions often require detailed
examination of the code. This suggests a need for the more abstract level
of description which dynamic grammars provide.

\subsection*{\small {\bf DYNAMIC GRAMMARS}}

Dynamics is just the study of states and transitions between states.
It can be used to specify the states of a left to right parser and the
possible mappings between states. For example, Milward (1994)
provides a dynamic description of a shift reduce parser, and a dynamic
description of a fully incremental parser based on dependency grammar.
Suitable languages for dynamics are both formal and declarative, and
are therefore also appropriate to express linguistic generalisations.

In a Dynamic Grammar (Milward 1994), each word is regarded as an
action which performs some change in the syntactic and semantic
context. For example, a parse of the sentence {\it John likes Mary}
becomes a mapping between an initial state, c$_{i}$, through some
intermediate states, c$_{a}$, c$_{b}$ to a final state c$_{f}$ i.e.
\begin{quote}
c$_{i}$ $\stackrel{John}{\mapsto}$ c$_{a}$
$\stackrel{likes}{\mapsto}$ c$_{b}$
$\stackrel{Mary}{\mapsto}$ c$_{f}$
\end{quote}
If we use a dynamic grammar to describe a shift reduce parser,
states encode the current
stack configuration,
and are related by rules which correspond to shifting and reducing
\footnote{Shift corresponds to: L $\mapsto$ $<$X$>\bullet$L on input of
a word, W, where L is a variable standing for a list
of categories, `$\bullet$' is list concatenation,
and X is the category for W. Reduce corresponds to
$<$C$_{n}$ ... C$_{1}>\bullet$L $\mapsto$ $<$C$_{0}>\bullet$L on empty
input, where
C$_{0}$ $\rightarrow$ C$_{1}$ ... C$_{n}$ is a phrase structure rule of the
grammar.}. Since there are arbitrarily large numbers of different stack
configurations (the stack can be of arbitrary size), the dynamics for
shift reduce parsing involves the use of an infinite number of states.
It thus differs from, say ATNs (Woods 1973), which have a finite number
of states, augmented by an explicit recursion mechanism.

Dynamic grammars can be presented as rewrite grammars by using {\it
transition types} instead of the more usual S or NP. For example, to
get the parse above we need the lexical entries:\footnote{For example,
for the shift reduce parser, the word {\it John} would get the type, L
$\mapsto$ $<$np$>\bullet$L, corresponding to a shifting of the NP onto
the stack. The empty string gets the type, $<$C$_{n}$
... C$_{1}>\bullet$L $\mapsto$ $<$C$_{0}>\bullet$L where C$_{0}$
$\rightarrow$ C$_{1}$ ... C$_{n}$ is a rule of the grammar,
corresponding to reduction.}
\begin{quote}
John:c$_{i}$$\mapsto$c$_{a}$ \hspace{1mm}
likes:c$_{a}$$\mapsto$c$_{b}$ \hspace{1mm}
Mary:c$_{b}$$\mapsto$c$_{f}$
\end{quote}
and a single combination rule schema which states that,
\begin{quote}
For any C1, C2, C3,\\[1mm]
C1$\mapsto$C3 \hspace{2mm} $\rightarrow$ \hspace{2mm}
C1$\mapsto$C2 \hspace{2mm} C2$\mapsto$C3
\end{quote}
A string of words is a sentence if it has the type,
\begin{quote}
c$_{i}\mapsto$c$_{f}$
\end{quote}
where {\bf c$_{i}$} and {\bf c$_{f}$} are appropriate initial and final
states for a parse\footnote{For the shift reduce parser, the initial state
is the empty list, $<>$, the final state is $<${\bf s}$>$.}.

In a dynamic grammar, any substring of a sentence can
be assigned a type. For example, {\it likes} and {\it Mary} can be
combined to get the type {\bf c$_{a}$$\mapsto$c$_{f}$}.  Thus we have
an appropriate level to perform substring coordination.  Dynamic
grammars may be extended using the following
combination rule ({\it and} and {\it or} are
both given the special transition type {\bf CONJ}):
{\small
\begin{quote}
For any C1, C2,\\[1mm]
C1$\mapsto$C2 \hspace{2mm} $\rightarrow$ \hspace{2mm}
C1$\mapsto$C2 \hspace{1mm} {\small CONJ} \hspace{1mm} C1$\mapsto$C2
\end{quote}}
\noindent
Similar to SYSCONJ, this allows coordination when two conjuncts map
between the same pairs of states.  Processing is also similar, with
the encountering of a conjunction causing back-up to an earlier stage
in the parsing history. However, since there is no popping of a stack,
the full parsing history is available\footnote{Something parallel to
popping occurs only after a coordination.  However this is exactly
what is required since we do not want overlapping coordination as in
{\it The girl and the or
the boy and the adult came}.}.
For example, {\it Ben gave some books to Sue}
has the transitions:
\begin{quote}
\hspace*{-4mm}
c$_{i}$ $\stackrel{Ben}{\rightarrow}$ c$_{k}$
$\stackrel{gave}{\rightarrow}$ c$_{l}$
$\stackrel{some}{\rightarrow}$ c$_{m}$
$\stackrel{books}{\rightarrow}$ c$_{n}$
$\stackrel{to}{\rightarrow}$ c$_{o}$
$\stackrel{Sue}{\rightarrow}$ c$_{f}$
\end{quote}
we can then parse {\it papers to Joe} using the transitions:
\begin{quote}
c$_{m}$
$\stackrel{papers}{\rightarrow}$ c$_{n}$
$\stackrel{to}{\rightarrow}$ c$_{o}$
$\stackrel{Joe}{\rightarrow}$ c$_{f}$
\end{quote}
Since the final state c$_{f}$ matches the state immediately before the
conjunction, the two strings can combine. The resulting transition diagram
is as follows:
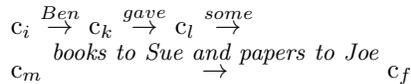
\begin{quote}
c$_{i}$ $\stackrel{Ben}{\rightarrow}$ c$_{k}$
$\stackrel{gave}{\rightarrow}$ c$_{l}$
$\stackrel{some}{\rightarrow}$ \\ c$_{m}$
$\stackrel{\mbox{\small{\it books to Sue and
papers to Joe}}}{\rightarrow}$ c$_{f}$
\end{quote}
Iterated coordination (e.g.\ for examples such as {\it Mary, Peter and
Sue}) can be treated in the same way as iterated
constituent coordination is treated in phrase structure grammars.
For example, each transition type can be augmented with a feature
(+/-) denoting whether or not that transition has been iterated.  The
coordination rule becomes:
{\small
\begin{quote}
\hspace*{-4mm}
For any C1, C2,\\[1mm]
\hspace*{-4mm}
C1$\mapsto^{-}$C2 \hspace{1mm} $\rightarrow$ \hspace{1mm}
C1$\mapsto^{+/-}$C2 \hspace{1mm} {\small CONJ} \hspace{1mm}
C1$\mapsto^{-}$C2
\end{quote}}
\noindent
Iterated types are formed as follows:
{\small
\begin{quote}
\hspace*{-4mm}
For any C1, C2,\\[1mm]
\hspace*{-4mm}
C1$\mapsto^{+}$C2 \hspace{2mm} $\rightarrow$ \hspace{2mm}
C1$\mapsto^{+/-}$C2 \hspace{2mm} C1$\mapsto^{-}$C2
\end{quote}}

The precise grammaticality predictions made by the dynamic approach
depend upon the characterisation of the states, and hence depend on
the particular parsing strategy which is specified by the dynamics.
However there are some general predictions which can be made.
Firstly, consider conjuncts which correspond one to one in the
categories of the corresponding words. Here the conjuncts must provide
the same transitions, and hence must be able to coordinate (this is a
reflection of the fact that processing can back up to any point in the
parsing history).  This predicts that any substring of a sentence can
coordinate with itself, and hence that any substring of a sentence can
act as a conjunct. For convenience we will call this the {\it
substring hypothesis}.  This hypothesis has been broadly adopted in
the work of van Oirsouw 1987, Barry and Pickering 1993, and by work on
the Lambek Calculus (e.g.\ Moortgat 1988).

Apparent counterexamples are as follows:
\begin{ex}
\begin{subex}
* The woman spoke to George and man to Peter
\end{subex}
\begin{subex}
* John told [Mary Bill] and [Fred Sue] was coming (Barry and Pickering 1993)
\end{subex}
\end{ex}
\mylabel{badsubstring}
However it is difficult to exclude these using syntactic
constraints, without
also excluding the more acceptable:
\begin{ex}
\begin{subex}
Every woman spoke to George and man spoke to Peter
\end{subex}
\begin{subex}
John told the mothers that their daughters and the fathers that their
sons were all at the party\footnote{This example is attributed by
Barry and Pickering (1993) to Janne Johannessen.}
\end{subex}
\end{ex}
\label{goodsubstring}
More natural examples where conjuncts are formed by fragments from different
constituents are the following:
\begin{ex}
\begin{subex}
The police found some [cars inside] and [lorries outside] the warehouse
\end{subex}
\begin{subex}
Everyone who I [admire most came] and [admire least stayed away]
\end{subex}
\begin{subex}
Mary showed many [friends the weird books] and [colleagues the more
respectable papers] written by her mother
\end{subex}
\end{ex}
\mylabel{police}
The relative unacceptability of the examples in \myref{badsubstring}{}
is perhaps best explained as due to violations of intonational
requirements, rather than syntactic requirements (cf. Steedman, 1989).

One case where the dynamic grammars correctly violate the substring
hypothesis is when a string already involves a coordination.
Here, the internal states are not accessible, so we can't get interleaving
of two coordinations, as in:
\begin{ex}
* The girl and the or
the boy and the adult came
\end{ex}
\mylabel{andor}
There may be an argument for similarly blocking coordination in
cases which would
involve the breaking apart of idioms or other structures which
are not standard cases of lexical subcategorisation.
An example (due to Mark Steedman), which may be such a case,
is the following,
\begin{ex}
* One man in [ten spoke against and twenty actually protested]
\end{ex}

As noted above, the precise grammaticality predictions depend on the
kind of parsing model which is encoded in the states.  In Milward
(1992), the dynamics specifies a word-by-word incremental parser for a
lexicalised version of dependency grammar.  Each state is a
recursively defined category, similar to a category in Categorial
Grammar.  For example, after parsing {\it You can call me} one
possible state is a sentence missing a sentence
modifier\footnote{Dependency grammar does not have VP modifiers}. This
state is appropriate as the initial state for a parse of both {\it
directly}, or of {\it after 3pm through my secretary}, resulting in a
final state of category sentence. Thus examples such as
\myref{secret}{a} are dealt with, since the syntactic context after
{\it You can call me} does not distinguish between one or more than
one subsequent modifier. This lack of distinction as to whether one or
more modifier is expected is actually a necessary prerequisite for
performing decidable fully word-by-word incremental interpretation
(see Milward and Cooper, 1994, in these proceedings).

Some of the problems with categorial grammar accounts of coordination do
reoccur with a dynamic account based on the parser used in Milward (1992).
For example,
\begin{ex}
{[}John] and [Mary thought that Peter] slept
\end{ex}
is predicted to be acceptable, as are the following,
\begin{ex}
\begin{subex}
{[}Today John] and [Mary thought that Peter] slept
\end{subex}
\begin{subex}
I heard [that] and [that no-one else knew that] Fred won the scholarship
\end{subex}
\end{ex}
This second batch of examples is particularly difficult to exclude
without making changes to the characterisation of the states. A
feature plus or minus tensed verb on each conjunct does block
them, but is difficult to motivate.

Dynamic grammars can be regarded purely as formal systems, as direct
representations of processing, or as something inbetween (for example,
in the packed parallel parser described in Milward (1994), the actual
parsing states are packed versions of the states in the grammar).
If we consider the dynamics to be a direct representation of processing,
then a dependence of linguistic data upon parsing states would only seem
plausible if the parsing process corresponds, at least to some extent, with
actual human language processing.
This brings up the
intriguing possibility that we can predict coordination facts from
known processing data, and vice versa.  For example, consider the well
known example of garden pathing:
\begin{ex}
The horse raced past the barn fell
\end{ex}
The choice between the use of {\it raced} as the main verb, or as part
of the reduced relative is usually assumed to be within the fragment
{\it the horse raced}, suggesting that there are two distinguished parsing
states after {\it raced}. Thus this correctly predicts the
unacceptability of the following:
\begin{ex}
{*} The horse raced [past the barn fell] and [beside the hedge]
\end{ex}

\subsection*{\small {\bf CONCLUSION}}

This paper has sketched various problems with some of the linguistic
accounts of coordination. It suggested that this was primarily due to
difficulty in encoding a proper notion of syntactic context. The paper
then considered various processing accounts, where the syntactic
context is encoded within the state of the parser.  Finally it showed
how dynamics can be used as a formal description of processing
accounts which use a full parsing history, and how the
characterisations of parsing states can be chosen to enforce the
requisite degree of parallelism between conjuncts.

\subsection*{\small {\bf REFERENCES}}

\noindent
Barry, G.\ and M.~Pickering (1993).
Dependency Categorial Grammar and Coordination.
{\it Linguistics, 31(5)}, p.855-902.\\[1.3mm]
 Dahl, V.\ and M.C.~McCord (1983). Treating Coordination in Logic Grammars.
{\it Computational Linguistics, 9-2}, p.69-91.\\[1.3mm]
Dowty, D. (1988). Type Raising, Functional Composition and Non-Constituent
Conjunction. In R.~Oehrle et al. Eds., {\it Categorial Grammars and
Natural Language Structures}. D.Reidel. \\[1.3mm]
Gleitman, L.R. (1965). Coordinating Conjunctions in English.
{\it Language, 41}, p.260-293. \\[1.3mm]
Goodall, G. (1987). {\it Parallel Structures in Syntax: Coordination
Causatives and Restructuring}, Cambridge University Press.\\[1.3mm]
Haugeneder, H. (1992). A Computational Model for Processing Coordinate
Structures: Parsing Coordination without Grammar. In {Proceedings of
ECAI 92}. \\[1.3mm]
Jorgensen, H. and A.~Abeill\'{e} (1992). Coordination of ``Unlike'' Categories
in TAG. In {\it Proceedings of the 2nd TAG Workshop}, Philadelphia. \\[1.3mm]
Lakoff, G.\ and S.~Peters (1969). Phrasal Conjunction and
Symmetric Predicates. In B.~Reibel and S.~Schane, Eds.,
{\it Modern Studies in English}. Eaglewood Cliffs: Prentice-Hall.\\[1.3mm]
Lambek, J. (1958). The Mathematics of Sentence Structure. {\it American
Mathematical Monthly, 65}, p.154-170.\\[1.3mm]
Milward, D.R. (1992). Dynamics, Dependency Grammar and Incremental
Interpretation. In {\it Proceedings of COLING 92}, Nantes,
p.1095-1099.\\[1.3mm]
Milward, D.R. (1994). Dynamic Dependency Grammar. To appear in
{\it Linguistics and Philosophy, 17}, p.561-605. \\[1.3mm]
Milward, D.R. and R.~Cooper (1994). Incremental Interpretation:
Applications, Theory, and
Relationship to Dynamic Semantics. In {\it Proceedings of COLING 94}, Kyoto,
Japan.\\[1.3mm]
Moltmann, F. (1992). {\it Coordination and Comparatives}. Ph.D.\ dissertation,
MIT, Cambridge Ma.\\[1.3mm]
Moortgat, M. (1988). {\it Categorial Investigations: Logical and Linguistic
Aspects of the Lambek Calculus}, Dordrecht: Foris. \\[1.3mm]
Morrill, G. (1990). Grammar and Logical Types. In
{\it Proceedings of the 7th Amsterdam Colloquium}. ITLI,
University of Amsterdam.\\[1.3mm]
Muadz, H. (1991). {\it Coordinate Structure: A Planar Representation}.
Ph.D.\ dissertation, University of Arizona, Tucson.\\[1.3mm]
van Oirsouw, R.R. (1987). {\it The Syntax of Coordination}.
Croom-Helm.\\[1.3mm]
Ross, J.R. (1967). {\it Constraints on Variables in Syntax}, Ph.D.\
dissertation, MIT, Cambridge Ma.\\[1.3mm]
Sag, I.A., G.~Gazdar, T.~Wasow  and S.~Weisler (1985).
Coordination and How to Distinguish Categories.
{\it Natural Language and Linguistic Theory, 3}, p.117-171. \\[1.3mm]
Steedman, M.J. (1985). Dependency and Coordination in the Grammar of Dutch
and English. {\it Language, 61}, p.523-568.\\[1.3mm]
Steedman, M.J. (1989). Intonation and Syntax in Spoken Language Systems.
Technical report, MV-CIS-89-20, Dept. of Computer and Information Science,
University of Pennsylvania. \\[1.3mm]
Wood, M.M. (1988). {\it A Categorial Syntax for Coordinate Constructions}.
Ph.D.\ Thesis, University College London. Available as Technical Report,
UMCS-89-2-1, Dept. of Computer Science, University of
Manchester.\\[1.3mm]
Woods, W. (1973). An Experimental Parsing System for Transition Network
Grammars. In R.~Rustin, Ed., {\it Natural Language Processing}, Algorithmics
Press, New York.
}
\end{document}